%% file: example_paper.tex
\documentclass{article}

\usepackage{microtype}
\usepackage{graphicx}
\usepackage{booktabs} 

\usepackage{hyperref}

\usepackage[accepted]{icml2025}

\usepackage{amsmath}
\usepackage{amssymb}
\usepackage{mathtools}
\usepackage{amsthm}

\usepackage[capitalize,noabbrev]{cleveref}

\theoremstyle{plain}

\theoremstyle{definition}

\theoremstyle{remark}

\usepackage[textsize=tiny]{todonotes}

\usepackage{hyperref}
\usepackage{url}
\usepackage{multirow}
\usepackage{booktabs, amsmath, colortbl, xcolor, graphicx}
\usepackage{lscape} 
\usepackage{subcaption}

\usepackage{wrapfig}
\definecolor{Gray}{gray}{0.85}
\newcolumntype{g}{>{\columncolor{Gray}}c}
\definecolor{Grayer}{gray}{0.65}
\newcolumntype{q}{>{\columncolor{Grayer}}c}

\icmltitlerunning{Two-Stage Pretraining for Molecular Property Prediction in the Wild}

\input{math_command}

\begin{document}

\twocolumn[
\icmltitle{Two-Stage Pretraining for Molecular Property Prediction in the Wild}



\icmlsetsymbol{equal}{*}

\begin{icmlauthorlist}
\icmlauthor{Kevin Tirta Wijaya}{yyy}
\icmlauthor{Minghao Guo}{comp}
\icmlauthor{Michael Sun}{comp}
\icmlauthor{Hans-Peter Seidel}{yyy}
\icmlauthor{Wojciech Matusik}{comp}
\icmlauthor{Vahid Babaei}{yyy}
\end{icmlauthorlist}

\icmlaffiliation{yyy}{Max Planck Institute for Informatics}
\icmlaffiliation{comp}{Massachusetts Institute of Technology}

\icmlcorrespondingauthor{Kevin Tirta Wijaya}{kwijaya@mpi-inf.mpg.de}

\icmlkeywords{Machine Learning, ICML}

\vskip 0.3in
]



\printAffiliationsAndNotice{}  

\begin{abstract}
Molecular deep learning models have achieved remarkable success in property prediction, but they often require large amounts of labeled data. The challenge is that, in real-world applications, labels are extremely scarce, as obtaining them through laboratory experimentation is both expensive and time-consuming.
In this work, we introduce MoleVers, a versatile pretrained molecular model designed for various types of molecular property prediction \textit{in the wild}, i.e., where experimentally-validated labels are scarce.
MoleVers employs a two-stage pretraining strategy.
In the first stage, it learns molecular representations from unlabeled data through masked atom prediction and \textit{extreme} denoising, a novel task enabled by our newly introduced branching encoder architecture and dynamic noise scale sampling.
In the second stage, the model refines these representations through predictions of auxiliary properties derived from computational methods, such as the density functional theory or large language models.
Evaluation on 22 small, experimentally-validated datasets demonstrates that MoleVers achieves state-of-the-art performance, highlighting the effectiveness of its two-stage framework in producing generalizable molecular representations for diverse downstream properties.
\end{abstract}

\section{Introduction}
Experimental chemistry is the domain of small data.
Determining the property of a molecule requires carefully designed experiments, constrained by material costs, reaction time, and specialized infrastructure.
This inherent scarcity forces chemists to extract insights from limited data points using statistical models, computational tools, and expert intuition, which in turn can be used to predict the properties of a novel molecule.
With recent advances in deep learning, researchers have increasingly turned to deep neural networks for learning molecular property prediction directly from data \citep{yang2019analyzing_dmpnn, rong2020self_grover, fang2022geometry_chemRLGEM, zhou2023unimol}.
However, these models typically rely on large datasets, where thousands or more labeled molecules are available for training.
Contrasts this to real-world datasets; of the 1,644,390 assays in the ChemBL database \citep{zdrazil2024chembl}, only 6,113 assays (0.37\%) contain 100 or more labeled molecules.
This discrepancy poses a challenge: how can we adapt data-hungry deep learning models to real-world applications, where even 50 training labels are considered plenty?

One promising solution is through unsupervised pretraining on large, unlabeled molecular datasets, followed by finetuning on specific properties using few labeled molecules.
Recent studies have explored this approach \citep{liu2022pretraining_graphmvp, xia2023molebert, zhou2023unimol, yang2024mol_molae}, demonstrating improvements on standard molecular benchmarks.
However, these benchmarks, such as MoleculeNet \citep{wu2018moleculenet}, contain thousands of labeled molecules---far from the reailty of most chemistry datasets.
When evaluated on small datasets, these pretrained models often show a significant drop in performance, sometimes only marginally surpassing the baseline of simply predicting the mean values of the labels.
Clearly, we need to rethink pretraining strategies in order to develop molecular property prediction models that are effective in real-world, small-data regime.


In this work, we introduce \textit{MoleVers}, a versatile pretrained model designed for molecular property prediction in data-scarce scenarios.
MoleVers is pretrained in two stages to maximize its generalizability to various types of downstream properties.
In the first pretraining stage, we propose a joint masked atom prediction (MAP) and \textit{extreme} denoising enabled by a novel branching encoder architecture and dynamic scale sampling.
The \textit{extreme} denoising approach allows the model to learn from a more diverse set of non-equilibrium configurations by utilizing a wider range of noise scales compared with conventional coordinate-based denoising. 
In the second pretraining stage, we train MoleVers to predict auxiliary properties that are different from, but more straightforward to obtain than the downstream properties.
These auxiliary properties can be calculated using density functional theory (DFT) or large language models (LLMs).
We use DFT to calculate three auxiliary properties: HOMO, LUMO, and dipole moment.
For LLM-based auxiliary labels, we propose to use pairwise rankings of molecular properties instead of absolute values, as we find that the LLM is more reliable at predicting relative rankings than absolute values.


To evaluate MoleVers, we introduce a new benchmark, Molecular Property Prediction in the Wild (MPPW). 
This benchmark consists of 22 small datasets curated from the ChemBL database \citep{zdrazil2024chembl}.
These datasets, most of which containing 50 or fewer training labels, span a wide range of molecular properties from physical characteristics to biological activities.
We standardized the pretraining datasets and data splits to ensure fair comparisons between MoleVers and several state-of-the-art pretrained models.
Experimental results show that MoleVers outperforms all baselines in 18 out of the 22 assays and ranks second in the remaining four, while no baseline method consistently ranks in the top two.
Moreover, MoleVers achieves state-of-the-art performance on large datasets in the MoleculeNet benchmark \citep{wu2018moleculenet}, highlighting the effectiveness of our two-stage pretraining strategy.

In summary, our contributions are: (1) a two-stage pretraining framework with a novel \textit{extreme} denoising task enabled by our branching encoder and dynamic noise scale sampling in the first stage, and auxiliary predictions based on DFT or LLM in the second stage, (2) a novel branching encoder architecture that facilitates the \textit{extreme} denoising pretraining, (3) a novel approach to utilize LLM as a molecular property ranker for auxiliary labels generation, and (4) the MPPW benchmark, designed to reflect real-world data limitations.
Source code is available at \hyperlink{https://github.com/ktirta/MoleVers}{https://github.com/ktirta/MoleVers}.

\section{Related Work}
Deep learning-based molecular property prediction has demonstrated remarkable successes.
Early approaches use graph neural networks (GNNs) to learn molecular representations directly from molecular structures \citep{kipf2017semisupervised_gcn, hamilton2017inductive_graphsage, veličković2018graph_gan}.
GNNs typically learn molecular representations by updating the node (atom) and edge (bond) features through a series of message passing across neighboring atoms.
Recently, popular property prediction benchmarks such as MoleculeNet \citep{wu2018moleculenet} are dominated by transformer-based models \citep{luo2022one_transformerM, zhou2023unimol, yang2024mol_molae} that leverage self-attention mechanisms to learn long-range interactions between atoms in a molecule.

Parallel to advances in architecture, pretraining has emerged as an effective strategy to improve property prediction performance when labeled data is limited.
By pretraining on a large, unlabeled dataset, a model can learn robust and transferable molecular representations that generalize well to a variety of downstream tasks.
Various pretraining strategies have been proposed, including masked predictions \citep{wang2019smiles_smilesbert, xia2023molebert, zhou2023unimol, yang2024mol_molae} and contrastive learning \citep{liu2022pretraining_graphmvp, xia2023molebert, wang2022molecular_molclr}.
Additionally, denoising atom coordinates and pairwise distance between them \citep{zaidi2023pretraining_forcefield, zhou2023unimol, liumolecular_geossl} have been shown to lead to strong downstream performance.
Pretraining via denoising is equivalent to learning an approximate molecular force field \citep{zaidi2023pretraining_forcefield, liumolecular_geossl}, which could explain its effectiveness for improving downstream property prediction performance.

Our work is also related to the few-shot molecular property prediction. 
Previous studies in this area \citep{ju2023few_fewshot1, guo2021few_fewshot2, wang2021property_fewshot3} often formulate the few-shot prediction as an N-way K-shot classification problem, where N classes of molecules are sampled from a dataset, each with K examples.
As this formulation is not directly applicable to regression tasks, we focus our discussion in the following sections to studies that follow the pretraining-finetuning paradigm.

\section{Two-Stage Pretraining}
Our primary objective is to develop an accurate molecular property prediction model that works in extremely small data regimes. 
To address this challenge, we propose a two-stage pretraining framework specifically designed to improve the generalization capability of our model, MoleVers. 
This approach enables accurate property prediction with only a few downstream labels.
In the following subsections, we discuss the details of MoleVers including the proposed \textit{extreme} denoising pretraining, branching encoder architecture, and auxiliary label generation with density functional theory and large language models.

\begin{figure*}
    \centering
    \includegraphics[width=0.88\linewidth]{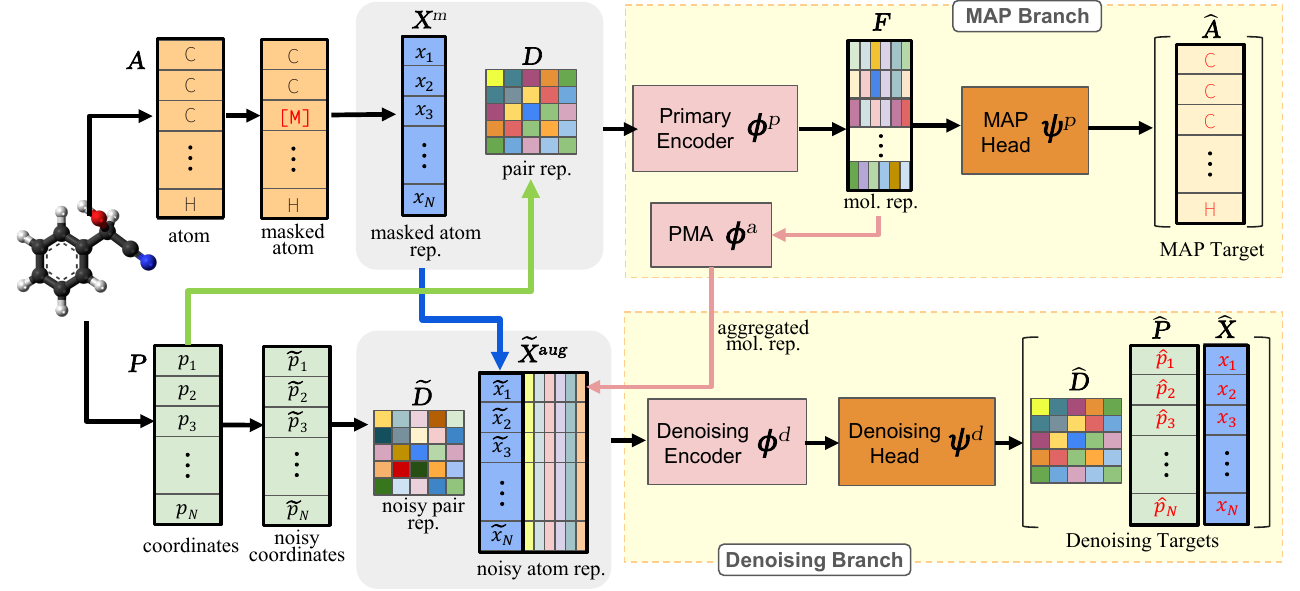}
    \caption{Illustration of pretraining \textbf{stage 1} using the proposed branching encoder. The primary encoder is assigned to the MAP branch, while another encoder with identical architecture is assigned to the denoising branch. For pretraining stage 2 and finetuning, we only keep the primary encoder and discard the denoising encoder.}
    \label{fig:branching_encoder_stage1}
\end{figure*}
\subsection{Stage 1: Masked Atom Prediction and Extreme Denoising}
\label{sec:map_dynamic_denoising}
In the first stage of our pretraining framework, we propose to train MoleVers on a large, unlabeled dataset using a combination of masked atom prediction (MAP) \citep{zhou2023unimol} and a novel \textit{extreme} denoising strategy, enabled by our branching encoder and dynamic noise scale sampling.
The following subsections provide a detailed discussion of these components.


\subsubsection{Masked Atom Prediction}
Inspired by masked token prediction in natural language processing (NLP) \citep{devlin2019bert, liu2019roberta, lewis2020bart}, masked atom prediction (MAP) involves training a model to predict the correct atom types in a partially-masked molecule.
This encourages the model to learn contextual relationship between atom types, capturing how they co-exist in various molecules.
Multiple works \citep{zhou2023unimol, xia2023molebert, yang2024mol_molae} have demonstrated the effectiveness of MAP as a pretraining task, which ultimately leads to better prediction models for the downstream datasets.
Following these studies, we use MAP as one of the two pretraining tasks in the first pretraining stage.

\subsubsection{Extreme Denoising}
\label{sec:dynamic_denoising}
To learn information from 3D structures, we employ coordinate and pairwise distance denoising.
\citet{zaidi2023pretraining_forcefield} and \citet{liumolecular_geossl} have shown that denoising tasks are equivalent to learning a molecular force field that is approximated with a mixture of Gaussians, $p(\Tilde{m}) \approx q_\sigma(\Tilde{m}):=\frac{1}{N}\sum_{i=1}^N q_\sigma(\Tilde{m} | m_i),$
where $p(\Tilde{m})$ is the force field, $q_\sigma(\Tilde{m} | m_i) = \mathcal{N}(\Tilde{m}; m_i, \sigma^2)$, and $m_1, m_2, ..., m_N$ are the equilibrium molecules in the pretraining dataset $\sD^{\text{train}}$.

Here, we propose \textit{extreme} denoising, a novel pretraining task that leverages relatively large noise scales to improve model generalization.
Our hypothesis is that using higher noise scale values  (e.g., $\sigma = 10$ instead of $\sigma = 1$) exposes the model to a broader set of non-equilibrium molecular configurations, ultimately improving its ability to generalize.
However, prior studies \citep{zhou2023unimol, yang2024mol_molae, ni2024pre_high_denoising} have reported that larger $\sigma$ values often deteriorate the downstream performance.
To address this challenge, we introduce two novel components: a branching encoder architecture which decouple the MAP and denoising pipelines, and the dynamic noise scale sampling.


\textbf{Decoupling MAP from Denoising.}
\label{sec:branching_encoder}
We first motivate the design choice of our branching encoder by examining the complexity differences of the MAP and denoising tasks.
In MAP, the model learns to map masked atoms ($\mA^\text{mask}$) to their corresponding atom logits ($\hat\mA$), $f(\mA^\text{mask}) = \hat\mA$, while in coordinate denoising, it learns to map noisy coordinates to their original values, $g(\Tilde{\mP}) = \hat\mP$.
The MAP function is relatively simpler because it maps a finite set of inputs (atom types) to a relatively compact set of outputs (softmax-normalized logits).
In contrast, denoising deals with continuous input and output coordinates, making it more complex as the number of possible mappings is much larger.
When a single model handles both MAP and denoising, the overall complexity is dominated by the more challenging denoising task.
The downstream performance could then be negatively affected if the model struggles to accurately fit the complex denoising function.

To mitigate this issue, we introduce a branching encoder architecture, shown in Figure \ref{fig:branching_encoder_stage1}, that decouples the MAP and denoising pipelines.
The branching design ensures that the complexity of the MAP task is only minimally affected by the denoising task.
Furthermore, we propose to connect the two encoders with an \textit{aggregator} module so that information can flow between the two pipelines.

\begin{figure*}[ht]
    \centering
    \includegraphics[width=0.9\linewidth]{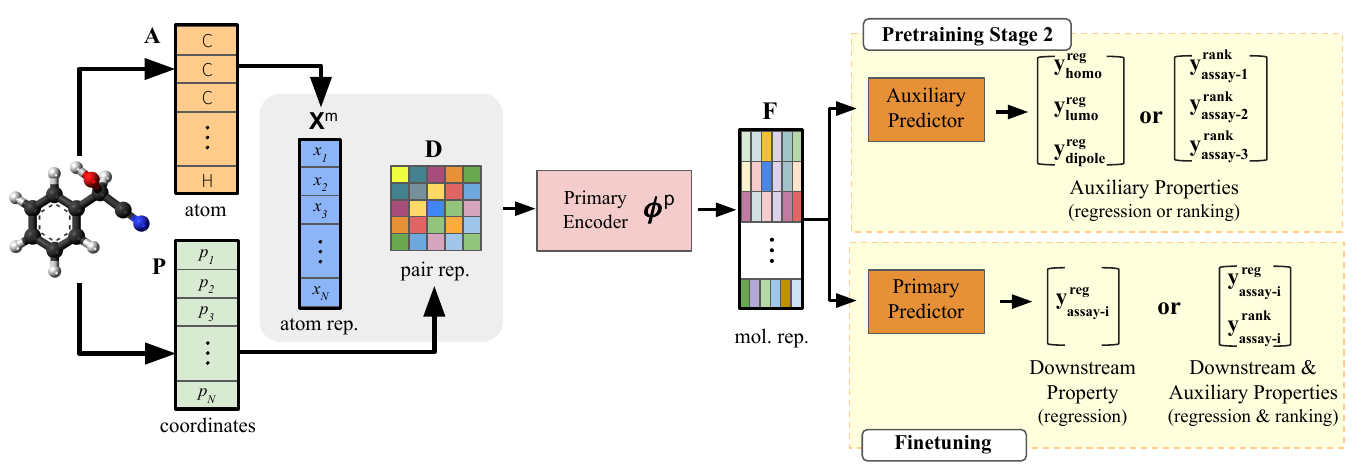}
    \caption{In pretraining \textbf{stage 2} and finetuning, we keep only the primary encoder to encode the atom and pair distance representations. A prediction head is appended to the model to predict the numerical properties of the molecules, denoted as \textbf{y}\textsuperscript{reg}. LLM-generated ranking labels, denoted as \textbf{y}\textsuperscript{rank}, can be used in the second pretraining stage or during finetuning.}
    \label{fig:stage2_finetuning}
\end{figure*}

\textbf{Branching Encoder.}
Inspired by prior work in NLP which have found masked prediction to often be the most effective pretraining tasks  \citep{lewis2020bart, raffel2020exploring_t5}, we set the MAP encoder as the primary encoder of the model.
The primary encoder, shown in Figure \ref{fig:branching_encoder_stage1}, will be further pretrained in the second pretraining stage (Figure \ref{fig:stage2_finetuning}) before being used for downstream predictions.

The branching encoder takes as input the types $\mA \in \sZ^{N}$ and coordinates $\mP \in \sR^{N \times 3}$ of the $N$ atoms in a molecule.
Following \citet{zhou2023unimol} and \citet{yang2024mol_molae}, each atom type is encoded into atom representation $\mX \in \sR^{N \times C}$ where $C$ is the number of features. The coordinates are transformed into pair distance representation $\mD \in \sR^{N \times N}$.
During the first pretraining stage, the atom representations are masked with a ratio of \textit{r}.
We denote the masked atom representations as $\mX^m$.

To extract the molecule representation $\mF$, we feed $\mX^m$ and $\mD$ into the primary MAP encoder $\phi^p$.
The logits $\hat\mA$ that represent the pristine atom types are then predicted with the MAP head $\psi^p$,
\begin{equation}
    \mF = \phi^p(\mX^m, \mD), \hspace{1.5em} \hat\mA = \psi^p(\mF).
\end{equation}

In the denoising branch, we inject noise sampled from a Gaussian distribution into $\mX$ and $\mP$ to obtain the noisy atom representations and coordinates,
\begin{equation}
    \Tilde{\mX} = \mX + \boldsymbol{\epsilon}_1, \hspace{1em} \Tilde{\mP} = \mP + \boldsymbol{\epsilon}_2,
\end{equation}
\begin{equation}
    \boldsymbol{\epsilon}_1 \sim \mathcal{N}(\textbf{0}, \sigma^2\mI_{N\times1}) , \hspace{0.5em} \boldsymbol{\epsilon}_2 \sim \mathcal{N}(\textbf{0}, \sigma^2\mI_{N\times3}), 
\end{equation}
To enable information flow from the denoising task to the primary MAP encoder, we augment $\Tilde{\mX}$ with an aggregated $\mF$ using a pooling with multihead attention (PMA) module \citep{lee2019set_pma},
\begin{equation}
    \Tilde{\mX}^{\text{aug}} = [\Tilde{\mX}, \phi^a(\mF)], \hspace{1em} \mG = \phi^d(\Tilde{\mX}^{\text{aug}}, \Tilde{\mD}, \sigma),
\end{equation}
\begin{equation}
     (\hat\mX, \hat\mP, \hat\mD) = \psi^d(\mG),
\end{equation}
where $\phi^d$ is the denoising encoder, $\phi^a$ is the PMA aggregator, $\Tilde{\mD}$ is derived from $\Tilde{\mP}$, and $\hat\mX, \hat\mP, \hat\mD$ are the denoising predictions of the denoising head $\psi^d$.


\textbf{Dynamic Noise Scale Sampling.} 
Conventional coordinate-based denoising typically employs a fixed noise scale $\sigma$ throughout the pretraining \citep{zaidi2023pretraining_forcefield, zhou2023unimol}.
However, we find that dynamically changing the noise scale during pretraining improves downstream performance.
Instead of using a static $\sigma$, we sample it from a uniform distribution, $\sigma \sim \mathcal{U}(0, a)$ where $a > 1$.
This approach effectively alters the variance of the Gaussian from which the noise is sampled, introducing a bias toward lower noise levels compared to a $\mathcal{N}(0, a)$ distribution (Figure \ref{fig:dynamic_static_noise}).
As a result, MoleVers is exposed more frequently to easier, low-noise samples while still ocassionaly encountering challenging, high-noise samples, leading to a more guided learning process.
In contrast, a static and low-valued $\sigma$ limits training to only easy samples, whereas a static and high-valued $\sigma$ makes difficult samples more frequent, potentially hindering the model's convergence.

\subsection{Stage 2: Auxiliary Property Prediction}
\label{sec:stage2}
We further improve the generalization capability of the primary encoder by incorporating auxiliary property prediction in the second pretraining stage.
This approach is inspired by multi-task learning \citep{caruana1997multitasklearning}, where a model is trained to solve both the primary task and related auxiliary tasks at the same time.
For example, in facial analysis, the primary task might be to predict facial landmarks, while the auxiliary tasks could be to estimate head poses and infer facial attributes \cite{zhang2014facial_mtl}.
Since these tasks share common features, the model can use the training signals from the auxiliary tasks to improve its performance in the primary task.

Given that molecular properties are heavily influenced by molecular structure, it is reasonable to assume that representations useful for predicting one type of property could also help in predicting others. 
Based on this intuition, we propose to construct an auxiliary dataset of properties that can be computed using relatively inexpensive computational methods, but are not necessarily identical to the properties in the downstream datasets. 
Specifically, we select highest occupied molecular orbital (HOMO), lowest unoccupied molecular orbital (LUMO), and dipole moment as the auxiliary properties because they can be accurately computed using density functional theory (DFT).
We also note that computing the auxiliary labels with DFT is significantly cheaper than obtaining more downstream labels via real-world experiments.

In this second pretraining stage, the model is trained in a supervised manner,
\begin{equation}
    \mF = \phi^p(\mX, \mD), \hspace{1.5em} (\hat y_{\text{homo}}, \hat y_{\text{lumo}}, \hat y_{\text{dipole}}) = \psi^q(\mF),
\end{equation}
where $\psi^q$ is the auxiliary predictor and $\hat y_{\text{homo}}, \hat y_{\text{lumo}}, \hat y_{\text{dipole}}$ are the predicted auxiliary properties.
Afterward, we append the primary predictor for the downstream property to the primary encoder and finetune the model using the downstream dataset, as illustrated in Figure \ref{fig:stage2_finetuning}.

\subsubsection{Auxiliary Label Generation with LLM}
Another promising approach in our second pretraining stage is to leverage large language models (LLMs) to generate the auxiliary labels.
LLMs have demonstrated strong generalization capability across a wide range of tasks due to their exposure to diverse pretraining data.
As a result, LLMs can be used to generate molecular property labels directly within the downstream task space with a fraction of computational cost of DFT.

Directly predicting the absolute values of properties using LLMs, however, often results in lower-quality labels.
Instead of absolute values, we propose to generate relative-valued labels in the form of pairwise rankings of molecules.
Pairwise rankings reformulate the problem as a binary classification task, where the input consists of two molecules and the output is either 0 or 1, indicating which molecule has a higher value for the given property.
These outputs can then be used either as auxiliary labels in the second pretraining stage, or as additional targets during finetuning (Figure \ref{fig:stage2_finetuning}).
We provide a more thorough discussion of LLM-based label generation in Appendix \ref{app:llm_label}.

\section{Molecular Property Prediction in the Wild Benchmark}
\label{sec:mppw}
The majority of existing molecular property prediction benchmarks rely on datasets with large numbers of data points, which do not reflect real-world scenarios where such large datasets are rare.
For instance, out of 1,644,390 assays available in the ChemBL database, only 6,113 assays (0.37\%) contain 100 or more molecules, demonstrating the scarcity of molecular data \textit{in the wild} where the molecular properties are validated through real-world experiments.
As a result, molecular property prediction models that perform well on existing benchmark may struggle to maintain the same level of performance in real-world applications where labeled data is limited.

To address this issue, we introduce Molecular Property Prediction in the Wild (MPPW), a new benchmark specifically designed for property prediction in low-data regimes.
Unlike existing benchmarks that often assume the availability of large and labeled datasets, the majority of datasets in the MPPW benchmark contain 50 or fewer training samples.
This reflects the challenge faced by molecular property prediction models \textit{in the wild}.
Specifically, we have curated 22 assays from the ChemBL database \citep{zdrazil2024chembl} that encompass a diverse set of properties that includes physical properties, toxicity, and biological activity.
A detailed description of the datasets, including their soruces, can be found in Appendix \ref{sec:benchmark_datasets}.

\section{Experiments and Results}
In this section, we address the following questions through a series of experiments: 
(1) Does the two-stage pretraining framework, including LLM-based generated labels, improve the downstream performance on datasets with significantly limited labels? 
(2) How does each individual pretraining stage contribute to the improvements? 
(3) Is our assumption that larger noise scales improve the generalization capability of the model correct?
(4) Does the choice of pretraining dataset affect downstream performance?
Additionally, we investigate how significant the impact of finetuning dataset size is to the downstream performance, for which the results are shown in the appendix.

\begin{figure*}[h]
    \centering
    \begin{subfigure}[b]{0.495\linewidth}
        \centering
        \includegraphics[width=\linewidth]{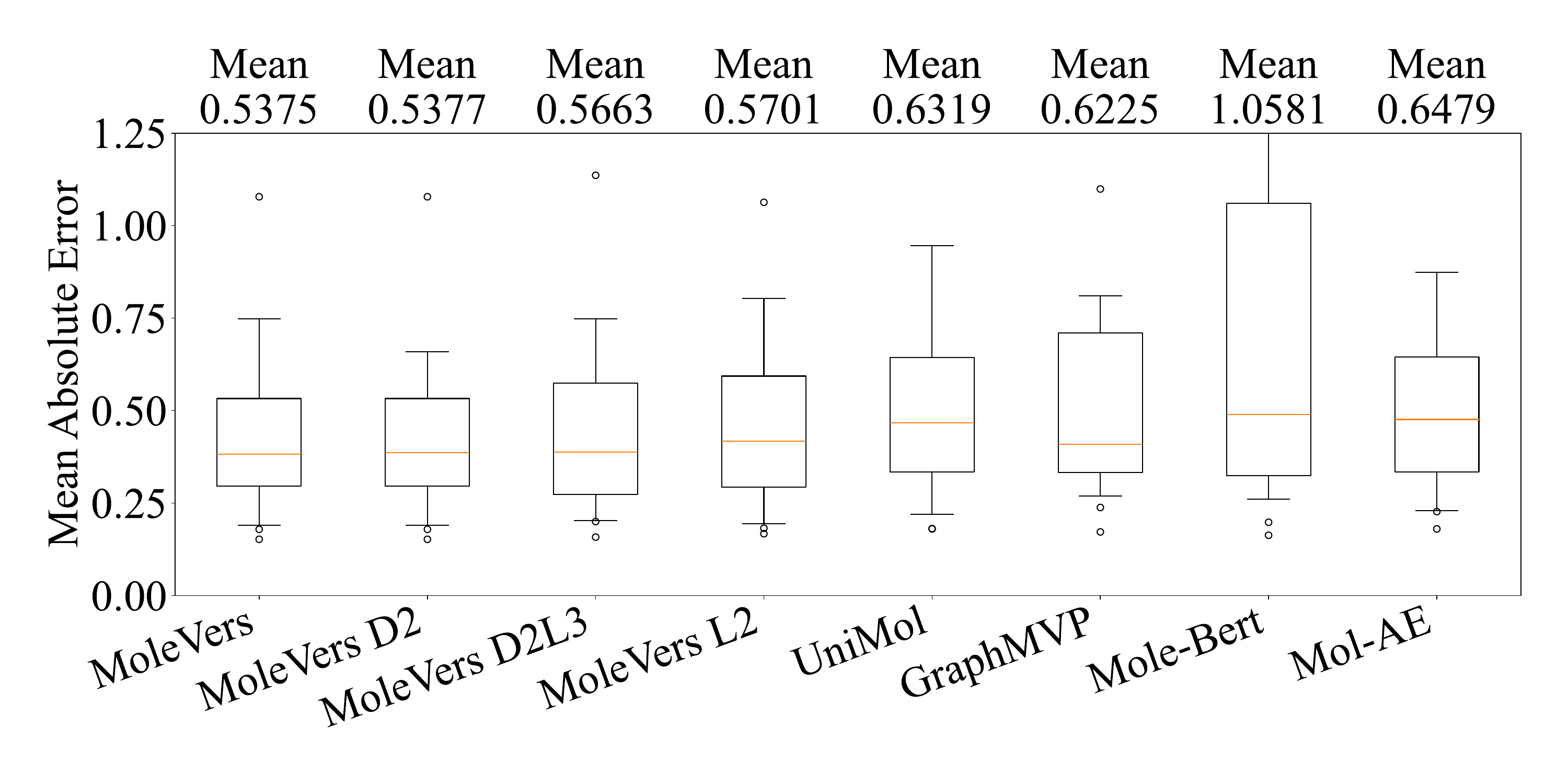} 
        \\ (a)
    \end{subfigure}
    \hfill
    \begin{subfigure}[b]{0.495\linewidth}
        \centering
        \includegraphics[width=\linewidth]{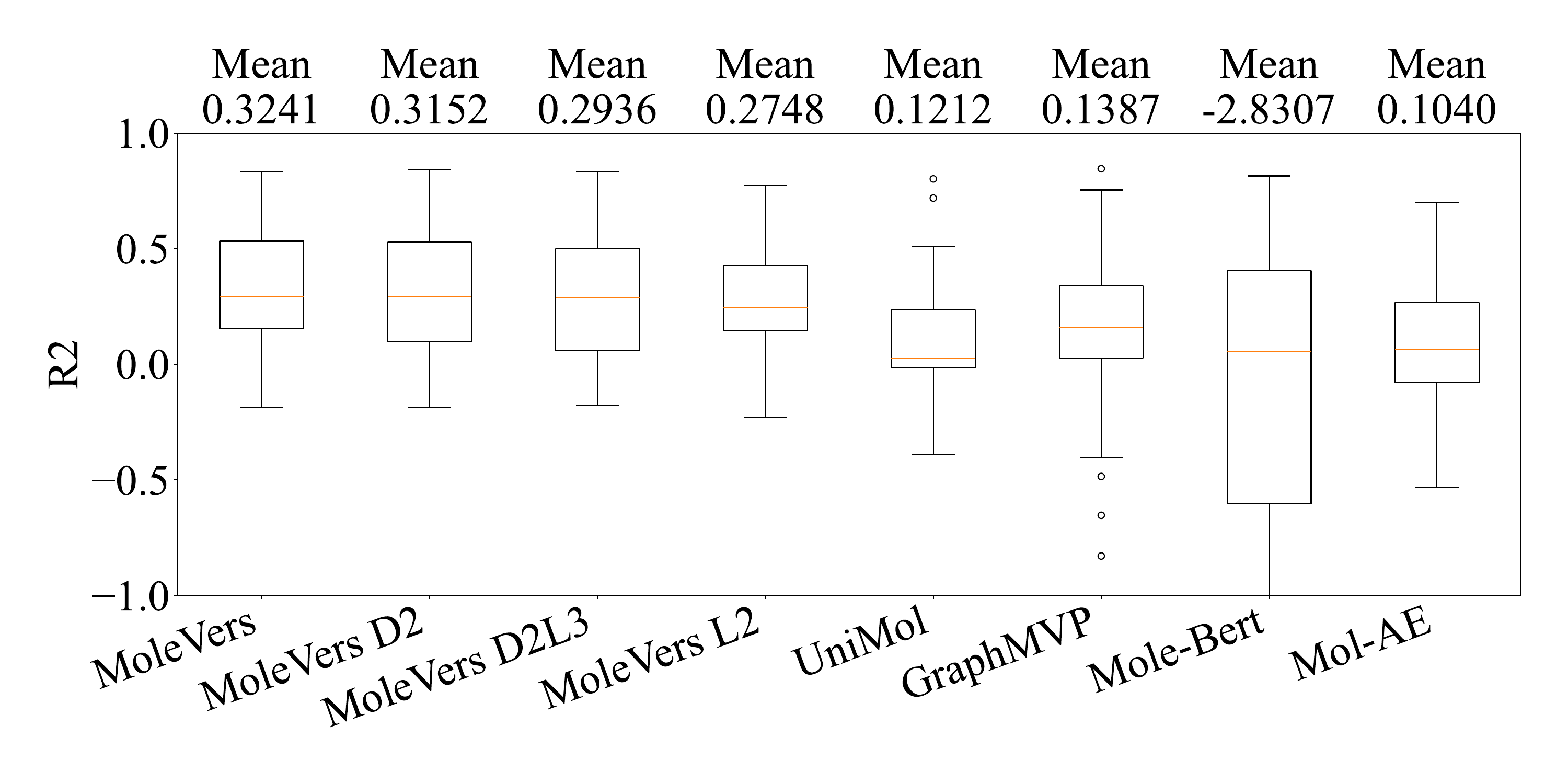} 
        \\(b)
    \end{subfigure}
    \caption{Box plot of (a) mean absolute error ($\downarrow$) and (b) R\textsuperscript{2} score ($\uparrow$) on the 22 assays and 3 train/test splits in the MPPW benchmark. The whiskers extend from the box to the farthest data point lying within 1.5x the inter-quartile range from the box. We can see that MoleVers variants, especially MoleVers, achieve the best overall performance with significant margins in both metrics. The full results are shown in Table \ref{tab:main_result_mppw} in Appendix.}
    \label{fig:mae_r2}
\end{figure*}
\begin{figure*}
    \centering
    \includegraphics[width=0.93\linewidth]{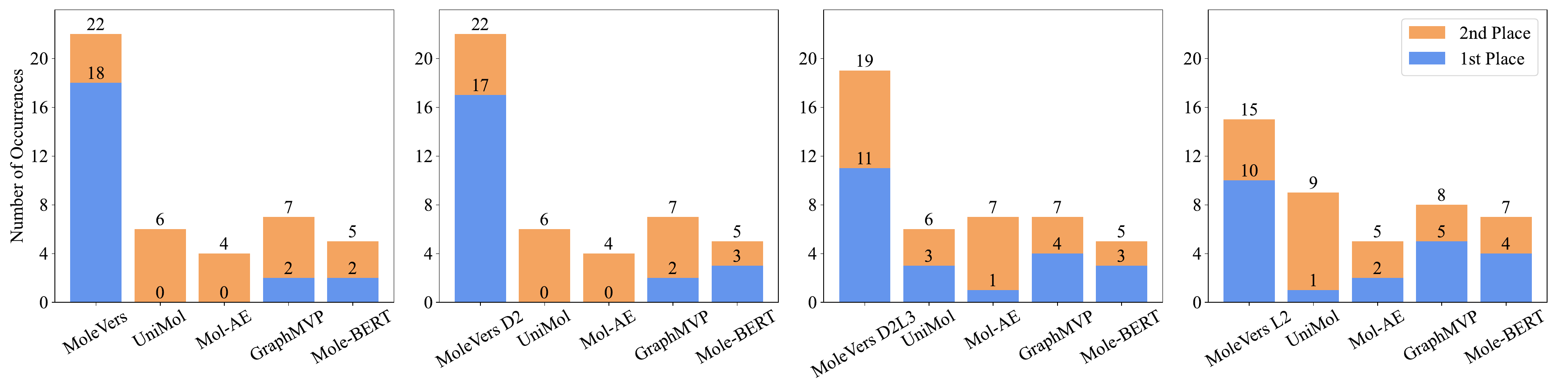}
    \caption{Number of assays in the MPPW benchmark (22 assays, 3 train/test splits) in which a model achieves the best or second best performance in terms of mean absolute error. Molevers demonstrates state-of-the-art performance, outperforming the baselines in 18 assays and achieving the second-best performance in the remaining 4 assays. }
    \label{fig:main_ranks}
\end{figure*}
 \subsection{Experiment Settings}
\textbf{Datasets.} We use GDB17 \citep{ruddigkeit2012enumeration_gdb17} as the pretraining dataset for our model \textit{and} other baselines to minimize any performance differences that might arise from the use of higher-quality pretraining datasets.
We randomly select 1M unlabeled molecules from the 50M subset to be used in the first pretraining stage.
We then sample 130K molecules out of the 1M subset to construct the auxiliary datasets for the second pretraining stage.
The labels for the DFT-based auxiliary dataset are computed with Psi4 \citep{smith2020psi4}, while ChatGPT4o is used for the LLM-based auxiliary dataset.
We use \citet{RDKit} to generate 3D conformations from SMILES \citep{weininger1988smiles} for models that take 3D graphs as input.

\textbf{Models \& baselines.} 
We provide evaluation results of four baselines: state-of-the-art GNNs, GraphMVP \citep{liu2022pretraining_graphmvp} and Mole-BERT \citep{xia2023molebert}, as well as state-of-the-art transformers, Uni-Mol \citep{zhou2023unimol} and Mol-AE \citep{yang2024mol_molae}.
All models are implemented in PyTorch \citep{paszke2019pytorch} and trained from scratch using publicly available source code.
Afterwards, the models are finetuned for 50 epochs on the \textit{downstream train split}, and the \textit{downstream test split} performance of the last epoch is recorded in Table \ref{tab:main_result_mppw}.
There are three train/test splits for each downstream dataset in the benchmark with a 1:1 train-test ratio.
We also provide comparisons with more baselines on large downstream datasets in Appendix \ref{sec:results_moleculenet}.
More details about network architecture and training strategy are available in Appendix \ref{app:network_training_details}.

We evaluated four variants of our proposed model, MoleVers, on the proposed MPPW benchmark. These four variants are different in second pretraining stage and finetuning strategies:
(1) \textbf{MoleVersD2}, pretrained on 130K labels generated by density functional theory (DFT) during the second stage,
(2) \textbf{MoleVersL2}, pretrained on 1M pairs of molecule ranking with labels generated by a large language model (LLM) during the second stage,
(3) \textbf{MoleVersD2L3}, pretrained on 130K DFT-generated labels during the second stage, then additionally finetuned using LLM-based ranking labels of the downstream test-split molecules, 
(4) \textbf{MoleVers}, pretrained on 130K DFT-generated labels. For this latest variant, LLM-generated ranking labels are only used if the absolute value of the rank correlation of the LLM predictions on the downstream training split is greater than 0.4.
Note that we finetune all variants on downstream train labels.

\textbf{Evaluation metrics.} We use two metrics for evaluation: mean absolute error (MAE) and the coefficient of determination (R\textsuperscript{2}).
Models with perfect predictions would achieve R\textsuperscript{2} scores of 1, while simply predicting the mean of the labels would result in an R\textsuperscript{2} score of 0.

\subsection{Results on the MPPW Benchmark}

\label{sec:main_results}
The predictive performance of the baseline models, evaluated on the MPPW benchmark, is presented in Figure \ref{fig:mae_r2} (and extensively in Table \ref{tab:main_result_mppw} in Appendix). 
From Figure~\ref{fig:mae_r2}, the poor R\textsuperscript{2} scores of the baselines, all below 0.14, indicate that their predictions are only marginally better than simply predicting the mean values of the properties. 
This highlights the need for more effective pretraining methods suited to small-data regimes. 
Note that Mole-BERT performs poorly on a few datasets, resulting in a significantly low average R\textsuperscript{2} score.

All variants of MoleVers outperform the baselines in terms of both average MAE and R\textsuperscript{2} scores. 
Figure \ref{fig:main_ranks} shows that MoleVers L2, pretrained exclusively with LLM-generated auxiliary labels, surpasses all baselines in 10 out of the 22 assays, and ranks second in 5 of the remaining assays. 
It also achieves improvements of 8.42\% in average MAE and 97.84\% in average R\textsuperscript{2} score compared to the best-performing baseline. 
This proves that a general-purpose LLM can be leveraged to generate weak labels for molecular property predictions. 
Such an approach can be useful in scenarios where access to more accurate tools, such as DFT, is limited due to resource or time constraints.

However, the availability of LLM-generated auxiliary labels does not always yield the optimal performance, particularly if more accurate auxiliary labels are accessible. 
For example, MoleVers D2, pretrained using DFT-based auxiliary labels, outperforms MoleVers D2L3, which incorporates LLM-based labels during finetuning. 
Nonetheless, MoleVers D2L3 still surpasses all baselines, outperforming them in 11 out of the 22 assays and ranking second in 8 of the remaining assays.

Interestingly, we observed that incorporating LLM-based labels during finetuning can improve performance in some assays. 
We hypothesize that the improvement correlates with the quality of the LLM predictions for a given assay. 
To test this hypothesis, we incorporate LLM-based labels during finetuning only when the absolute rank correlation coefficient of the LLM predictions on the training set is above 0.4. 
This strategy results in our main model, MoleVers, which outperforms the baselines in 18 out of the 22 assays and ranks second in the remaining 4 assays. 
MoleVers demonstrates improvements of 13.67\% in average MAE and 133.09\% in average R\textsuperscript{2} score compared to the best-performing baseline. 
Overall, these results confirm that the two-stage pretraining framework is an effective approach for improving downstream performance when labeled data is extremely limited.

\subsection{Ablation of Pretraining Stages}
We study the influence of each pretraining stage on the downstream performance of MoleVers through a series of ablation studies.
We use MoleVers D2 in all of our ablation studies.
As shown in Table \ref{tab:results_ablation_stages}, incorporating either the first or second pretraining stage into the pipeline always leads to better downstream performance compared with directly training the model on the downstream datasets.
Interestingly, the improvements vary across assays: some benefit more from the first pretraining stage, while others see more gains from the second pretraining stage.
This variation could be due to the auxiliary properties we have chosen--HOMO, LUMO, and Dipole Moment--which are more related to intrinsic molecular properties (e.g., assay 1), rather than complex interactions (e.g., assay 3).
Overall, the combination of both pretraining stages consistently yields the best downstream performance across all assays.

\begin{table}[t]
    \small
    \centering
    \caption{Ablation studies of our pretraining strategy. We report the mean MAE ($\downarrow$) across three train/test splits. We can see that combining both pretraining stage 1 and stage 2 gives the best performance on the downstream datasets.}
    \begin{tabular}{cccccc}
    \toprule
    Pretrain & Pretrain & \multicolumn{4}{c}{Assay ID} \\
    \cmidrule(lr){3-6}
    Stage 1& Stage 2 & 1 & 2 & 3 & 4 \\
    \cmidrule(lr){1-6} 
    - & - & 0.683 &  0.493 & 3.680 & 0.784 \\
     \checkmark & - & 0.592 & 0.420 & 3.161 & 0.431 \\
     - & \checkmark & 0.501 & 0.343 & 3.301 & 0.346 \\
     \checkmark & \checkmark & \textbf{0.417} & \textbf{0.298} & \textbf{2.999} & \textbf{0.337} \\
    \bottomrule
    \end{tabular}
    \label{tab:results_ablation_stages}
\end{table}

\subsection{Ablation of Branching Encoder and Extreme Denoising}
The key components that enable denoising pretraining with higher noise levels in the first stage are the branching encoder and dynamic noise scale sampling.
Here, we study the impact of each component to the downstream performance.
As shown in Table \ref{tab:results_ablation_be_dd}, using a single encoder for denoising pretraining at larger noise levels generally leads to worse prediction performance.
In contrast, the introduction of the branching encoder can mitigate this issue in most cases.
Furthermore, combining the branching encoder with dynamic noise scale sampling consistently yields the best downstream performace, highlighting the importance of dynamic over static noise scales.

\begin{table}[t]
    \small
    \centering
    \caption{Ablation studies of the proposed branching encoder and \textit{extreme} denoising. B.E. and D.D. stands for branching encoder and dynamic denoising, respectively. We report the mean MAE ($\downarrow$) across 3 train/test splits. Combining branching encoder with varying noise levels of a broader range yields the best downstream performance.}
    \begin{tabular}{ccccccc}
    \toprule
    B.E. & D.D. & Max $\sigma$ & \multicolumn{4}{c}{Assay ID} \\
    \cmidrule(lr){4-7}
    &  &  & 1 & 2 & 3 & 4 \\
    \cmidrule(lr){1-7} 
    - & - & 1 &  0.481 & 0.426 & 3.393 & 0.475 \\
    - & - & 10 &  0.519 & 0.418 & 3.401 & 0.492 \\
     \checkmark & - & 10 & 0.521 & 0.336 & 3.301 & 0.476 \\
     \checkmark & \checkmark & 10  & \textbf{0.417} & \textbf{0.298} & \textbf{2.999} & \textbf{0.337} \\
    \bottomrule
    \end{tabular}
    \label{tab:results_ablation_be_dd}
\end{table}

\begin{table}[t]
    \small
    \centering
    \caption{Effects of noise scales on downstream performance. We report the mean MAE ($\downarrow$) across three train/test splits. Larger noise scales tend to improve the downstream performance of MoleVers. However, using excessively large noise scales (e.g., max. $\sigma$ = 20) leads to unstable training.}
    \begin{tabular}{rcccc}
        \toprule
        \multicolumn{1}{c}{Max. Noise} & \multicolumn{4}{c}{Assay ID} \\
        \cmidrule(lr){2-5}
         Scale $\sigma$ & 1 & 2 & 3 & 4 \\
         \cmidrule(lr){1-5}
        0.1     & 0.944 & 0.414 & 3.321 & 0.443\\
        1       & 0.658 & 0.464 & 3.486 & 0.559 \\
        3       & 0.592 & 0.420 & 3.161 & 0.431 \\
        10      & \textbf{0.417} & \textbf{0.298} & \textbf{2.999} & \textbf{0.337} \\
        20      & - & - & - & - \\
         \bottomrule
    \end{tabular}
    \label{tab:result_ablation_noisescale}
\end{table}

\subsection{Impact of Noise Scale on Downstream Performance}
In Section \ref{sec:dynamic_denoising}, we hypothesized that using larger noise scales for the denoising tasks can improve the downstream performance.
In Table \ref{tab:result_ablation_noisescale}, we show the downstream performance of MoleVers with various noise scales.
Note that, similar to what has been observed in a prior work \citep{yang2024mol_molae}, the pretraining become unstable when excessively larger noise scales, e.g., $b = 20$, are used.
Therefore, we limit our ablation to a maximum value of 10.

We can see from Table \ref{tab:result_ablation_noisescale} that, as the maximum noise scale increases, we observe consistent improvements in performance.
The results confirm our hypothesis that larger noise scales could improve the downstream performance if implemented carefully.
This also highlights the importance of the proposed branching encoder, which facilitates denoising pretraining with larger noise scales.

\subsection{Impact of Pretraining Dataset Quality on Downstream Performance}
In Section \ref{sec:mppw}, we hypothesized that much of the performance gains observed in previous works may stem more from the quality of the pretraining datasets than from the pretraining method itself.
Therefore, it is important to fix the pretraining dataset used in a benchmark.
To test this, we examine two factors: the size of the pretraining dataset and its molecular diversity.
Intuitively, a larger and more diverse set of pretraining molecules should lead to a better pretrained model compared to smaller pretraining datasets with less variation.

Table \ref{tab:results_pretraining_dataset_size} shows the downstream performance of MoleVers when pretrained on datasets of varying sizes in the first stage.
We observe a general trend of improved downstream performance as the pretraining dataset size increases.
One exception occurs in Assay 2, where the model pretrained on 100K samples outperforms the one pretrained on 1M samples.
However, the R\textsuperscript{2} difference between these two models is relatively small compared to other assays, therefore, the overall trend remains valid.
Furthermore, we investigate the impact of pretraining dataset diversity by filtering out molecules containing specific atom types.
As shown in Table \ref{tab:results_pretraining_dataset_atomtypes}, downstream performance generally improves as the molecular diversity of the pretraining dataset increases.

These results confirm that large and diverse pretraining datasets can improve molecular property on downstream datasets.
They also highlight the importance of standardizing pretraining datasets when comparing different pretraining methods.
Specifically, using the same pretraining datasets, as was done in the MPPW benchmark, ensures that any observed downstream performance improvements are the results of the pretraining strategy itself rather than variations in the pretraining dataset quality.

\begin{table}[t]
    \small
    \centering
    \caption{Impact of dataset diversity (stage 1), measured by the number of training samples. We report the mean MAE ($\downarrow$) across 3 train/test splits. The performance of MoleVers improves as the number of training samples increases.}
    \begin{tabular}{rcccc}
        \toprule
        \multicolumn{1}{c}{Training size} & \multicolumn{4}{c}{Assay ID} \\
        \cmidrule(lr){2-5}
          & 1 & 2 & 3 & 4 \\
         \cmidrule(lr){1-5}
        10,000   & 1.152 & 0.498 & 3.660 & 0.611 \\
        100,000  & 0.629 & \textbf{0.409} & 3.205 & 0.549\\
        1,000,000 & \textbf{0.592} & 0.420 & \textbf{3.161} & \textbf{0.431} \\
         \bottomrule
    \end{tabular}

    \label{tab:results_pretraining_dataset_size}
\end{table}

\begin{table}[t]
    \small
    \centering
    \caption{Impacts of pretraining (stage 1) dataset diversity, measured by the variety of atom types. The number of molecules in each dataset is fixed to 100K for a fair comparison. We report the mean MAE ($\downarrow$) across 3 train/test splits. M stands for miscellaneous atom types. The downstream performance of MoleVers improves when the number of unique atom types in the training set increases.}
    
    \begin{tabular}{ccccccccc}
        \toprule
        \multicolumn{5}{c}{Atom Types} & \multicolumn{4}{c}{Assay ID} \\
        \cmidrule(lr){6-9}
          & & & & & 1 & 2 & 3 & 4 \\
         C & N & O & F & M. & \\
         \cmidrule(lr){1-9}
               
        \checkmark & \checkmark & - & - & - & 1.093 & 0.496 & 3.584 & 0.628 \\

        \checkmark & \checkmark & \checkmark & - & - & 0.845 & 0.431 & 3.428 & 0.480 \\

        \checkmark & \checkmark & \checkmark & \checkmark & - & 0.619 & 0.423 & 3.273& 0.493 \\
        
        \checkmark & \checkmark & \checkmark & \checkmark & \checkmark & \textbf{0.592} & \textbf{0.420} & \textbf{3.161} & \textbf{0.431}\\
         \bottomrule
    \end{tabular}
    \label{tab:results_pretraining_dataset_atomtypes}
\end{table}

\section{Conclusion}
We addressed the challenge of molecular property prediction \textit{in the wild}---scenarios where experimentally-validated labels are significantly limited---using a two-stage pretraining strategy.
In the first stage, we introduced a novel \textit{extreme} denoising pretraining task, enabled by our branching encoder and dynamic noise scale sampling.
In the second stage, we pretrained the model with auxiliary labels obtained through high-accuracy methods such as density functional theory, or more computationally efficient approaches such as large language models.
We evaluated our model on a new benchmark, Molecular Property Prediction in the Wild, with extremely small datasets that were chosen to reflect real-world data limitations.
Our model consistently outperforms existing baselines, making it suitable for real-world applications where labeled data are limited.

\newpage
\bibliography{example_paper}
\bibliographystyle{icml2025}

\newpage
\appendix
\onecolumn
\section{Appendix}
\subsection{Leveraging Large Language Models for Auxiliary Labels Generation}
\label{app:llm_label}
In this section, we discuss the use of large language models for generating auxiliary molecular property labels.
In our experiments, we employ the ChatGPT 4o model as the label generator.
The model is initialized with the prompt provided in Table \ref{tab:llm_input}, and a CSV file containing pairs of molecules along with the target property of interest are uploaded to the model.
The predictions are saved in a downloadable CSV file.
We have encountered instances where the model either declines to provide outputs or returns placeholder values (e.g., all 0s or 1s).
In such cases, restarting the process from the initial prompt typically resolves the issue.
\begin{table}[h]
    \centering
    \caption{Initial prompt to the LLM.}
\noindent\fbox{%
    \parbox{\textwidth}{%
You will be given a csv file that consists of pairs of molecules represented as SMILES strings. The user will input the molecular property of interest in the same prompt. Your task is to (1) explain briefly what does the property mean, (2) explain briefly in a paragraph what makes a molecule exhibit a higher value in said property, (3) based on your explanation and existing knowledge, for each pair, rank the molecules. Output 0 if p\textsubscript{smiles1} $>$ p\textsubscript{smiles2}, and 1 otherwise, where p\textsubscript{smiles} is the property of the molecule. Do not use any external tools. You should output your rank predictions as a csv file, where each line consists of 'smiles1, smiles2, prediction'.}%
}
    \label{tab:llm_input}
\end{table}

There are two ways to utilize LLM-generated auxiliary labels: during the second pretraining stage or during finetuning. 
In MoleVers L2, we employ the LLM to predict pairwise rankings for 1 M pairs of molecules sampled from the GDB17 dataset. 
The molecular properties of the first seven assays in the MPPW benchmark serve as the target properties.
In MoleVers D2L3, LLM predictions are incorporated during finetuning. 
First, for one assay, the LLM predicts pairwise rankings for all molecules in the test set. 
Then, the regression labels from the training split and the auxiliary ranking labels from the test split are used to finetune the model.

\subsubsection{LLM as property ranker or regressor}
The prompt in Table \ref{tab:llm_input} can also be adapted to generate regression predictions, i.e., the absolute property values of each molecule.
However, our experiments indicate that ChatGPT 4o performs better at predicting pairwise rankings than absolute property values.
Table \ref{tab:llm_ranking_vs_regression} compares the predictive performance of ChatGPT 4o for regression and ranking tasks across the first 13 assays in the MPPW benchmark.
R$^2$ and Kendall's $\tau$ are employed as the evaluation metrics for regression and ranking, respectively.
An ideal regression prediction would result in an $R^2$ score of 1, whereas simply predicting the mean would yield a score of 0.
Similarly, a perfect ranking would correspond to a $|\tau|$ value of 1, while a completely randomized ranking would score 0.
As shown in the table, ChatGPT consistently achieves equal or better performance in predicting pairwise rankings compared to absolute property values.
Therefore, we use pairwise rankings as the auxiliary labels in our main experiments rather than absolute property values.
\begin{table}[ht]
    \centering
    \caption{Comparison of ChatGPT's predictive performance for regression (absolute property values) and pairwise ranking (relative property values). We can see that ChatGPT consistently outputs higher-quality ranking predictions compared with absolute value predictions.}
    \begin{tabular}{lccccccccccccc}
    \toprule
         Assay ID & 1 & 2 & 3 & 4 & 5 & 6 & 7 & 8 & 9 & 10 & 11 & 12 & 13\\
         Regression ($R^2$) & 0.50 & 0.22 & 0.00 & 0.37 & 0.00 & 0.01 & 0.15 & 0.10 & 0.05 & 0.05 & 0.02 & 0.14 & 0.00\\
         Ranking ($|\tau|$) & 0.50 & 0.34 & 0.04 & 0.44 & 0.06 & 0.08 & 0.17 & 0.11 & 0.08 & 0.24 & 0.05 & 0.23 & 0.13\\
    \bottomrule
    \end{tabular}

    \label{tab:llm_ranking_vs_regression}
\end{table}

\subsubsection{Effects of LLM prediction quality to the downstream prediction performance}
The performance of LLM ranking predictions can be evaluated using at least three metrics: accuracy, Kendall's $\tau$, and the absolute value of Kendall's $\tau$ ($|\tau|$). 
From our experiments, we observe weak correlations between MAE improvements and the first two metrics: -0.00557 for LLM accuracy and -0.00274 for LLM's $\tau$. 
However, a much stronger correlation of -0.30563 is observed between MAE improvements and $|\tau|$. 
This indicates that there is a correlation between the quality of LLM ranking predictions and the downstream prediction performance (higher $|\tau|$ values leads to reduction in MAE). 

It is noteworthy that this relationship is not evident when using accuracy or raw $\tau$ as metrics. 
This can be explained by the nature of the pairwise ranking, a binary classification task, employed during training with auxiliary labels.
In this setting, LLM-predictions of very low accuracy (e.g., 0\%) or $\tau$ score (e.g., -1) provides signals that are equally strong to very high accuracy (e.g., 100\%) or $\tau$ score (e.g., +1).

To illustrate this, consider a set of LLM-generated labels with an accuracy of 50\%. 
Such a set is equivalent to random noise, offering no useful information. 
However, a set of labels with 0\% accuracy, while seemingly poor by conventional metrics, contains perfect (although inverted) information, as every prediction is consistently wrong. 
Similarly, a $\tau$ score of -1 indicates a perfect reverse ranking, holding the same signal strength as a $\tau$ score of +1.
For this reason, the absolute value of Kendall's $\tau$ ($|\tau|$) is a more reliable predictor of downstream MAE improvements compared to raw accuracy or $\tau$ scores. 

\subsubsection{In-context learning for LLM ranking}
One approach to improve a large language model on a given task is by providing it with relevant examples, a process also known as in-context learning.
Here, we explore the possibility of adapting ChatGPT 4o to our assays through in-context learning.
For a given assay, we provide the model with molecules from the training split, along with their corresponding property labels.
The model is then tasked with generating the auxiliary ranking labels for the test molecules.

\begin{figure}
    \centering
    \includegraphics[width=0.8\linewidth]{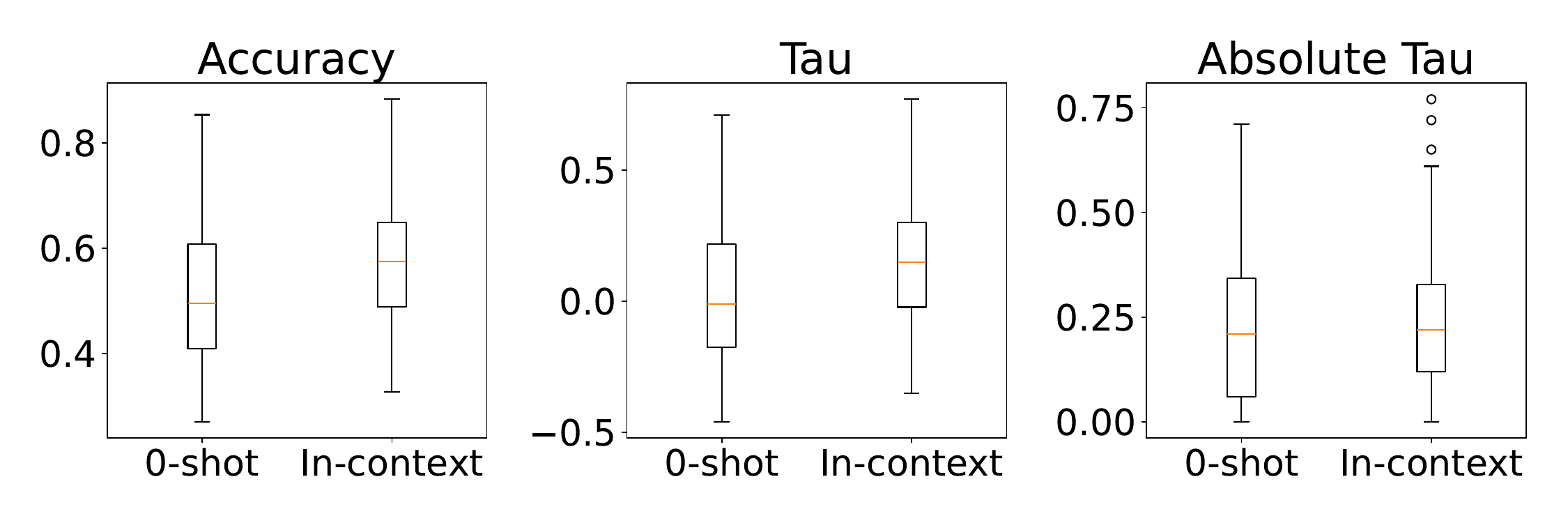}
    \caption{Predictive performance comparison between ChatGPT 0-shot and ChatGPT with in-context learning across 22 assays in the MPPW benchmark. While in-context learning improves the predictive accuracy and Kendall's tau in general, the absolute value of Kendall's tau remains relatively unchanged.}
    \label{fig:boxplot_0shot_incontext}
\end{figure}

As shown in the box plots in Figure \ref{fig:boxplot_0shot_incontext}, incorporating additional data generally improves both accuracy and $\tau$ values. 
However, the absolute $\tau$ values remain relatively unchanged. 
This is because the observed improvements primarily stem from re-aligning the LLM predictions to the pre-determined labeling convention. 
For example, ChatGPT might yield a 0-shot $\tau$ value of -0.26 and an in-context $\tau$ value of 0.09. 
In this scenario, the 0-shot model tends to predict flipped rankings correctly, while the in-context model more accurately predicts the actual rankings. 
Although the in-context approach shifts the $\tau$ value in a positive direction, the magnitude is deacreasing. 
As a result, the in-context model shows improvement in overall $\tau$, but not in absolute $|\tau|$.

\subsection{Network and Training Details}
\label{app:network_training_details}
The primary and auxiliary encoders of MoleVers are built on the UniMol encoder architecture \citep{zhou2023unimol}.
Each encoder comprises 15 layers, with an embedding dimension of 512 and a feedforward dimension of 2048.
The MAP and denoising heads are implemented with multilayer perceptrons, while the aggregator module is implemented with pooling by multihead attention (PMA), a cross attention-based module introduced by \citet{lee2019set_pma}.
The PMA module uses a \textit{query} of size $1 \times 512$, and takes the molecule features $F$ as \textit{key} and \textit{value}.
During the first pretraining stage, the model is trained for 1 million iterations using a batch size of 32, with a masking ratio of 0.15 for the MAP task.
In the second pretraining stage, the model is trained for 50 epoch, maintaining the same batch size of 32.
We employ the Adam optimizer with a learning rate of 10\textsuperscript{-4} and utilize a polynomial decay learning rate scheduler.
We run all experiments on an NVIDIA Quadro RTX 8000 GPU.

\subsection{Details of datasets used in of the MPPW Benchmark}
\label{sec:benchmark_datasets}

The Molecular Property Prediction in the Wild (MPPW) benchmark uses two types of datasets: pretraining datasets and downstream datasets. 
For our first-stage pretraining, as well as in the pretraining of other models shown in Table \ref{tab:main_result_mppw}, we randomly select 1M unlabeled molecules from the GDB17 dataset \citep{ruddigkeit2012enumeration_gdb17}. 
For the second-stage pretraining, we sample around 130K molecules from the 1M subset and calculate the auxiliary labels—HOMO, LUMO, and Dipole Moment—using Psi4 \citep{smith2020psi4}. 

For downstream evaluation, we curated 22 small datasets from the ChemBL database \citep{zdrazil2024chembl}, representing a diverse set of molecular properties as detailed in Table \ref{tab:app_dataset_details}. 
To ensure consistency across datasets, we filter out any molecules containing atoms not present in the GDB17 dataset. 
As a result, only molecules containing the atoms \{H, C, N, O, S, F, Cl, Br, I\} are included in the downstream datasets.
For evaluation, each dataset is randomly sampled to create three train/test splits with a 50:50 ratio, and all models in Tables \ref{tab:main_result_mppw} are assessed using these same splits.
The processed datasets can be accessed through this \href{https://drive.google.com/file/d/1VXaiOf6X4t9GxYU2G6NQqxRppIBTdMB3/view?usp=sharing}{URL}.

\begin{table}[ht]
    \footnotesize
    \centering
    \caption{Details for datasets in the MPPW benchmark. We curated 22 small datasets of diverse properties from the ChemBL database. The last two datasets are used for ablation (Section \ref{sec:finetuning_size}).}
    \begin{tabular}{rrrlll}
    \toprule
        ID & {ChemBL ID} & \#Mols. & \multicolumn{1}{c}{Short Description} & \multicolumn{1}{c}{Unit} & \multicolumn{1}{c}{Reference} \\
        \cmidrule(lr){1-6}
        1   & 635482    & 100 & Partition coefficient (logP)          & - & \citet{hansch1980antitumor_assay1}\\
        2   & 4150258   & 99 & Antimycobacterial activity against    & log nM & \citet{nyantakyi2018indolyl_assay2}\\
            &           & & Mycobacterium bovis BCG ATCC          & & \\
            &           & & 35734  & & \\
        3   & 744489    & 94 & Antimalarial activity in Plasmodium   & - & \citet{kesten1992synthesis_assay3}\\
            &           & & berghei infected mice (Mus musculus)  & & \\
        4   & 638473    & 48 & Partition coefficient (logD7.4)       & - & \citet{rai1998synthesis_assay4}\\
        5   & 5251479   & 51 & Induction of mitochondrial            & log nM & \citet{murray2023oxadiazolopyridine_assay5}\\ 
            &           & & uncoupling activity in rat L6 cells   \\
            &           & & assessed as increase in oxygen    \\
            &           & & consumption rate \\
        6   & 778368    & 95 & Hypolipidemic effects(plasma TG) in  & \% & \citet{sircar1983phenylenebis_assay6} \\
            &           & & male rats \\
        7   & 813331    & 69 & Inhibitory activity against   & log nM & \citet{vedani2000multiple_assay7}\\
            &           & & Tachykinin receptor 1 \\
        8   & 3375151   & 60 & Antimycobacterial activity against  & log nM & \citet{karabanovich20141_assay8}\\
            &           & & Mycobacterium kansasii CNCTC My    \\
            &           & & 235/80 \\
        9   & 687437    & 68 & Bronchodilator activity against & log umol  & \citet{hermecz1987nitrogen_assay9}\\
            &           & & histamine- induced spasm in & kg\textsuperscript{-1} \\
            &           & & guinea pig \\
        10  & 4770530   & 78 & Cytotoxicity against human TZM-GFP  & log nM & \citet{wang2020chemical_assay10}\\
            &           & & cells \\
        11  & 3282634   & 75 & Antitumor activity against mouse L1210  & log mg & \citet{denny1978potential_assay11}\\
            &           & & cells transfected in ip dosed C3H/DBA2  & kg\textsuperscript{-1} day\textsuperscript{-1}  \\
            &           & & F1 mouse qd \\
        12  & 632430    & 50 & Partition coefficient (logP) (chloroform) & - & \citet{dunn1987role_assay12}\\
        13  & 950577    & 44 & Antifungal activity against Candida  & - & \citet{katritzky2008qsar_assay13}\\
            &           & & albicans \\
        14  & 984427    & 85 & Antiviral activity against CVB2 infected  & log nM & \citet{tonelli2008antimicrobial_assay14}\\
            &           & & in Vero76 cells \\
        15  & 1862759   & 96 & DNDI: Lipophilicity measured in  & - & \\
            &           & & Chromatographic hydrophobicity index \\
            &           & & assay, pH 7.4 \\
        16  & 3066822   & 47 & Dissociation constant, pKa of the  & - & \citet{akamatsu2011importance_assay16} \\
            &           & & compound at pH 7.3 \\
        17  & 3745095   & 84 & Antifungal activity against Candida  & log\textsubscript{2} ug & \citet{de2016anti_assay17} \\
            &           & & glabrata clinical isolate & ml\textsuperscript{-1} \\
        18  & 4835984   & 61 & Brain to blood partition coefficient of the  & - & \citet{li2021discovery_assay18}\\
            &           & & compound \\
        19  & 4888494   & 123 & Re-testing in dose-response curve in  & log nM & \citet{dechering2022replenishing_assay19}\\
            &           & & HepG2 cytotoxicity assay, at 72h \\
        20  & 5043600   & 101 & Cytotoxicity in dog MDCK cells assessed   & log nM & \citet{mizuta2021lead_assay20}\\
            &           & & as reduction in cell viability \\
        21  & 1070367   & 38 & ABTS radical scavenging activity assessed   & log MU & \citet{amic2010reliability_assay21}\\
            &           & & as trolox equivalent antioxidant capacity\\
        22  & 2427705   & 44 & Half life in phosphate buffer at pH 7.4   & log hour & \citet{ward2013structure_assay22}\\
            &           & & at 50 uM \\
        A & 5291763 & 1237 & Inhibition of NaV1.7 ion channel & log nM & \citep{sutherland2023preclinical_assayA}\\
        B & 2328568  & 1017 & Inhibition of human CHRM1 & log nM & \citep{norinder2013qsar_assayB}\\

    \bottomrule
    \end{tabular}
    \label{tab:app_dataset_details}
\end{table}

\subsection{Details of experimental results in the MPPW benchmark}
\begin{figure}[h]
    \centering
    \includegraphics[width=0.8\linewidth]{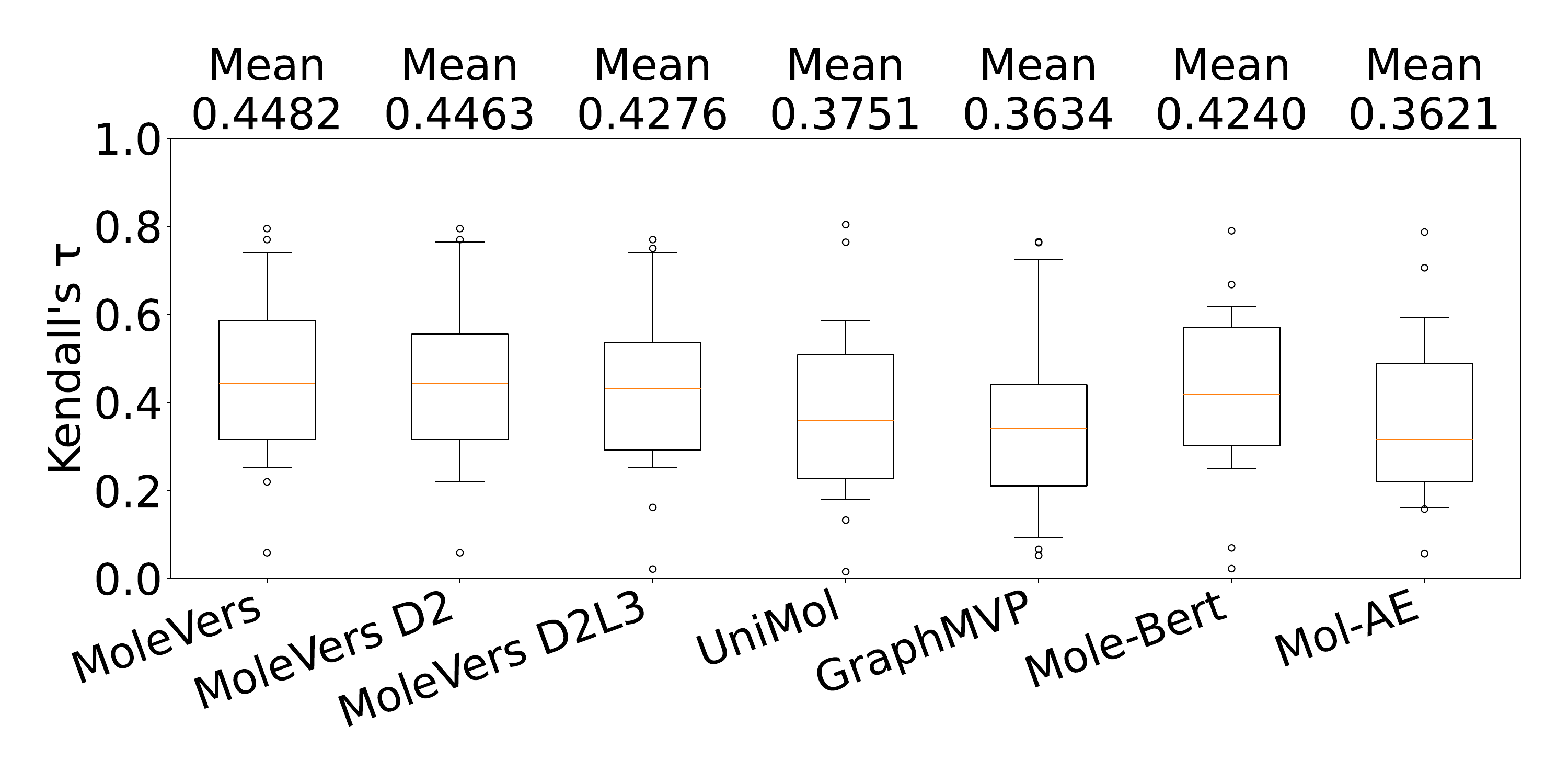}
    \caption{Box plot of Kendall's tau rank correlation coefficient (absolute values, $\uparrow$). The whiskers extend from the box to the farthest data point lying within 1.5x the inter-quartile range from the box.}  
    \label{fig:boxplot_tau}
\end{figure}

Table \ref{tab:main_result_mppw} presents the numerical values of the metrics discussed in Section \ref{sec:main_results}. 
In addition to the MAE and R\textsuperscript{2} scores, we report the Kendall rank correlation coefficient ($\tau$) for MoleVers and the baselines. 
Kendall's $\tau$ ranges from -1 to 1, where a perfect rank prediction yields +1, a random ranking results in 0, and a score of -1 indicates a completely inverted ranking relative to the ground truth.
The rank correlation coefficient is particularly relevant since performance optimization is a common objective in real-world applications
Identifying molecules with better properties (e.g., higher solubility, higher activity), is therefore as important as predicting the absolute property values. 
To illustrate the results, we provide box plots of $\tau$ scores in Figure \ref{fig:boxplot_tau}. 
Similar to the MAE and R\textsuperscript{2} results, MoleVers outperforms all baselines by a substantial margin.

\begin{table*}[t]
    \centering
    \caption{Quantitative results on the MPPW benchmark. We report the mean MAE ($\downarrow$), R\textsuperscript{2} ($\uparrow$), and Kendall's Tau-b across 3 train/test splits for each assay.}
    \begin{tabular}{ccgqcgqcgq}
    \toprule
        \multirow{2}{*}{Assay ID} & \multicolumn{3}{c}{\textbf{GraphMVP}} & \multicolumn{3}{c}{\textbf{MoleBert}}  & \multicolumn{3}{c}{\textbf{UniMol}} \\
        \cmidrule(lr){2-4} \cmidrule(lr){5-7} \cmidrule(lr){8-10} 
        & MAE & R$^2$ & $\tau$ & MAE & R$^2$ & $\tau$ & MAE & R$^2$ & $\tau$ \\
    \cmidrule(lr){1-10}
    1  & 0.426 & 0.846 & 0.763  & 0.526 & 0.724 & 0.668 & 0.459 & 0.802 & 0.764 \\
    2  & 0.408 & 0.260 & 0.398  & 0.303 & 0.528 & 0.617 & 0.455 & 0.067 & 0.267 \\
    3  & 3.459 & 0.024 & 0.229  & 3.351 & -0.011 & 0.296 & 3.282 & 0.035 & 0.252 \\
    4  & 0.329 & 0.716 & 0.725 & 0.429 & 0.500 & 0.619 & 0.522 & 0.186 & 0.573 \\
    5  & 0.463 & -0.201 & -0.093 & 2.277 & -19.732 & 0.251 & 0.415 & -0.129 & 0.391\\
    6  & 0.269 & -0.653 & 0.053 & 0.197 & -0.066 & 0.070 & 0.180 & -0.018 & 0.016 \\
    7  & 0.660 & 0.169 & 0.292 & 0.522 & 0.421 & 0.498 & 0.672 & -0.060 & 0.251 \\
    8  & 0.810 & -0.829 & -0.205 & 3.449 & -19.905 & 0.279 & 0.667 & 0.017 & 0.204 \\
    9  & 0.341 & 0.102 & 0.299 & 0.261 & 0.328 & 0.407 & 0.347 & -0.011 & 0.133 \\
    10 & 0.238 & 0.285 & 0.398 & 0.418 & -0.693 & 0.535 & 0.220 & 0.300 & 0.437 \\
    11  &0.409 & 0.178 & 0.347 & 0.456 & -0.107 & 0.258 & 0.428 & 0.019 & 0.362\\
    12  & 0.727 & 0.133 & 0.396 & 0.797 & 0.149 & 0.398 & 0.653 & 0.316 & 0.582 \\
    13  & 0.630 & 0.051 & 0.147 & 1.148 & -1.969 & 0.023 & 0.592 & 0.068 & 0.308 \\
    14  & 0.343 & 0.036 & 0.334 & 0.443 & -0.351 & 0.430 & 0.330 & 0.132 & 0.381 \\
    15  & 0.378 & 0.703 & 0.627 & 0.563 & 0.356 & 0.500 & 0.474 & 0.512 & 0.586 \\
    16  & 0.753 & 0.357 & 0.455 & 0.784 & 0.351 & 0.471 & 0.946 & -0.039 & 0.355 \\
    17  & 1.099 & 0.238 & 0.373 & 1.151 & 0.023 & 0.351 & 1.347 & -0.390 & 0.220\\
    18  & 0.172 & 0.754 & 0.765 & 0.163 & 0.815 & 0.790 & 0.181 & 0.719 & 0.804 \\
    19  &0.305 & -0.485 & -0.102 & 0.307 & -0.687 & 0.321 & 0.223 & 0.012 & 0.179 \\
    20  & 0.329 & -0.402 & 0.067 & 0.271 & 0.089 & 0.362 & 0.303 & 0.000 & 0.217 \\
    21 & 0.343 & 0.622 & 0.680 & 0.375 & 0.532 & 0.583 & 0.589 & -0.123 & 0.532  \\
    22 & 0.805 & 0.148 & 0.247 & 5.086 & -23.570 & 0.600 & 0.616 & 0.251 & 0.438 \\
    \end{tabular}
    
    \vspace{0.1cm} 

    \begin{tabular}{ccgqcgq}
    \toprule
        \multirow{2}{*}{Assay ID} & \multicolumn{3}{c}{\textbf{MolAE}} & \multicolumn{3}{c}{\textbf{MoleVers}} \\
        \cmidrule(lr){2-4} \cmidrule(lr){5-7} 
        & MAE & R$^2$ & $\tau$ & MAE & R$^2$ & $\tau$ \\
    \cmidrule(lr){1-7}
    1  & 0.607 & 0.698 & 0.787  & 0.393 & 0.832 & 0.770 \\
    2  & 0.356 & 0.305 & 0.502  & 0.298 & 0.516 & 0.554 \\
    3  & 3.470 & -0.096 & 0.239 & 2.999 & 0.211 & 0.370 \\
    4  & 0.475 & 0.399 & 0.706 & 0.332 & 0.701 & 0.740 \\
    5 & 0.402 & 0.012 & 0.320 & 0.401 & -0.134 & 0.336 \\
    6 & 0.180 & -0.016 & 0.057 & 0.179 & -0.016 & 0.059 \\
    7 & 0.742 & -0.169 & 0.158 & 0.650 & 0.018 & 0.252 \\
    8 & 0.630 & 0.079 & 0.249 & 0.593 & 0.144 & 0.310 \\
    9 & 0.351 & -0.042 & 0.194 & 0.339 & 0.034 & 0.220 \\
    10 & 0.286 & -0.089 & 0.313 & 0.190 & 0.457 & 0.481 \\
    11 & 0.444 & -0.110 & 0.312 & 0.372 & 0.302 & 0.432 \\
    12 & 0.649 & 0.425 & 0.560 & 0.536 & 0.538 & 0.598 \\
    13 & 0.611 & 0.156 & 0.358 & 0.522 & 0.181 & 0.335 \\
    14 & 0.328 & 0.093 & 0.368 & 0.295 & 0.287 & 0.454 \\
    15 & 0.477 & 0.522 & 0.572 & 0.447 & 0.591 & 0.546 \\
    16 & 0.874 & 0.058 & 0.451 & 0.748 & 0.278 & 0.557 \\
    17 & 1.451 & -0.533 & 0.162 & 1.078 & 0.192 & 0.254 \\
    18 & 0.227 & 0.572 & 0.592 & 0.152 & 0.822 & 0.795 \\
    19 & 0.230 & -0.130 & 0.243 & 0.220 & -0.188 & 0.365 \\
    20 & 0.296 & -0.046 & 0.203 & 0.235 & 0.347 & 0.220 \\
    21 & 0.513 & 0.132 & 0.407 & 0.326 & 0.648 & 0.684 \\
    22 & 0.654 & 0.068 & 0.214 & 0.519 & 0.369 & 0.463 \\
    
    \bottomrule
    \end{tabular}
    \label{tab:main_result_mppw}
\end{table*}

\subsection{Results on Large Downstream Datasets}
\label{sec:results_moleculenet}
We further evaluate the performance of MoleVers on the MoleculeNet benchmark \citep{wu2018moleculenet}, focusing on large-scale regression datasets such as QM7, QM8, and QM9.
As shown in Table \ref{tab:results_moleculenet}, MoleVers outperforms all baseline models across all datasets, achieving the lowest MAE scores.
Therefore, the proposed two-stage pretraining framework is not only effective in low-data regimes, but also excels when abundant labeled data is available.
We note, however, scenarios in which thousands of labeled data is available is extremely rare in the real-world.

\begin{table}[ht]
    \small
    \centering
    \caption{Results on larger datasets. We use three large regression datasets of the MolculeNet benchmark: QM7, QM8, and QM9. The MAE values of methods other than MoleVers are obtained from \citet{yang2024mol_molae}.}.
    \begin{tabular}{lccc}
    \toprule
         Dataset & QM7 & QM8 & QM9 \\
         \#Molecules & 6830 & 21789 & 133885 \\
    \cmidrule(lr){1-4}
        D-MPNN & 103.5 & 0.0190 & 0.0081 \\
        Attentive FP & 72.0 & 0.0179 & 0.0081 \\
        Pretrain-GNN & 113.2 & 0.0200 & 0.0092 \\
        GROVER & 94.5 & 0.0218 & 0.0099 \\
        MolCLR & 66.8 & 0.0178 & - \\
        Uni-Mol & 58.9 & \underline{0.0160} & 0.0054 \\
        Mol-AE & \underline{53.8} & 0.0161 & \underline{0.0053} \\
        \cmidrule(lr){1-4}
        MoleVers (ours) & \textbf{51.3} & \textbf{0.0155} & \textbf{0.0050} \\

    \bottomrule
    
    \end{tabular}
    \label{tab:results_moleculenet}
\end{table}

\subsection{Impact of Finetuning Dataset Size on Downstream Performance}
\label{sec:finetuning_size}
\begin{figure}[ht]
    \centering
    \includegraphics[width=0.8\linewidth]{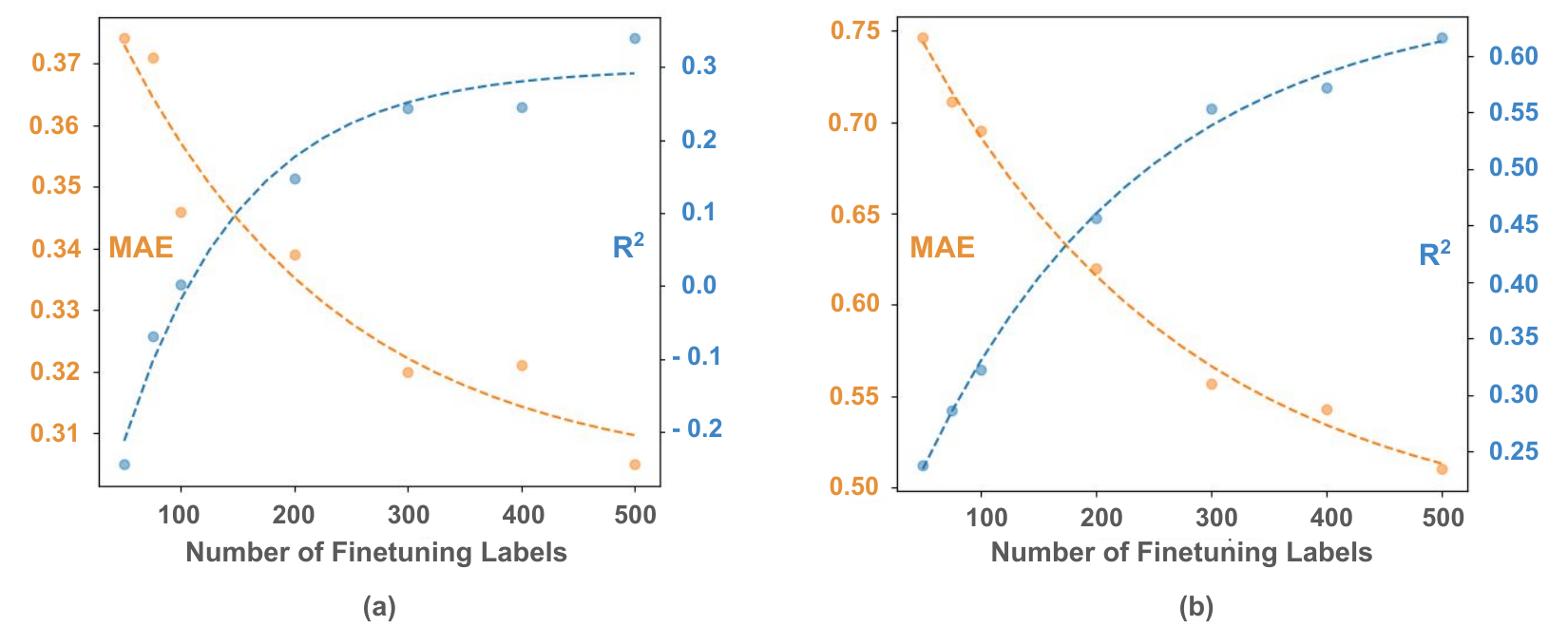}
    \caption{Predictive performance of MoleVers, averaged over 5 splits, when finetuned on two assays with varying dataset size: (a) CHEMBL5291763, (b) CHEMBL2328568 \citep{zdrazil2024chembl}.}
    \label{fig:finetuning_dataset_size}
\end{figure}

To assess the impact of finetuning dataset size on downstream performance, we gradually reduce the number of training labels used to finetune MoleVers, and validate it on fixed validation sets.
We conduct this experiment using two large datasets outside the MPPW benchmark, as the datasets in the benchmark contain only a limited number of molecules.
As shown in Figure \ref{fig:finetuning_dataset_size}, the MAE curves show exponential decay as the number of finetuning labels increases, while the R\textsuperscript{2} curves exhibit logarithmic growth.
This demonstrates a sharp drop in prediction quality, especially when the number of finetuning labels fall below 200.
These results emphasize the inherent challenge of molecular property prediction \textit{in the wild} due to the scarcity of labeled data in real-world.
The observed performance degradation with smaller datasets also highlights the importance of an effective pretraining strategy, such as the proposed two-stage pretraining approach of MoleVers, in mitigating the limitations imposed by limited labeled data.

\begin{figure}[h]
    \centering
    \includegraphics[width=0.8\linewidth]{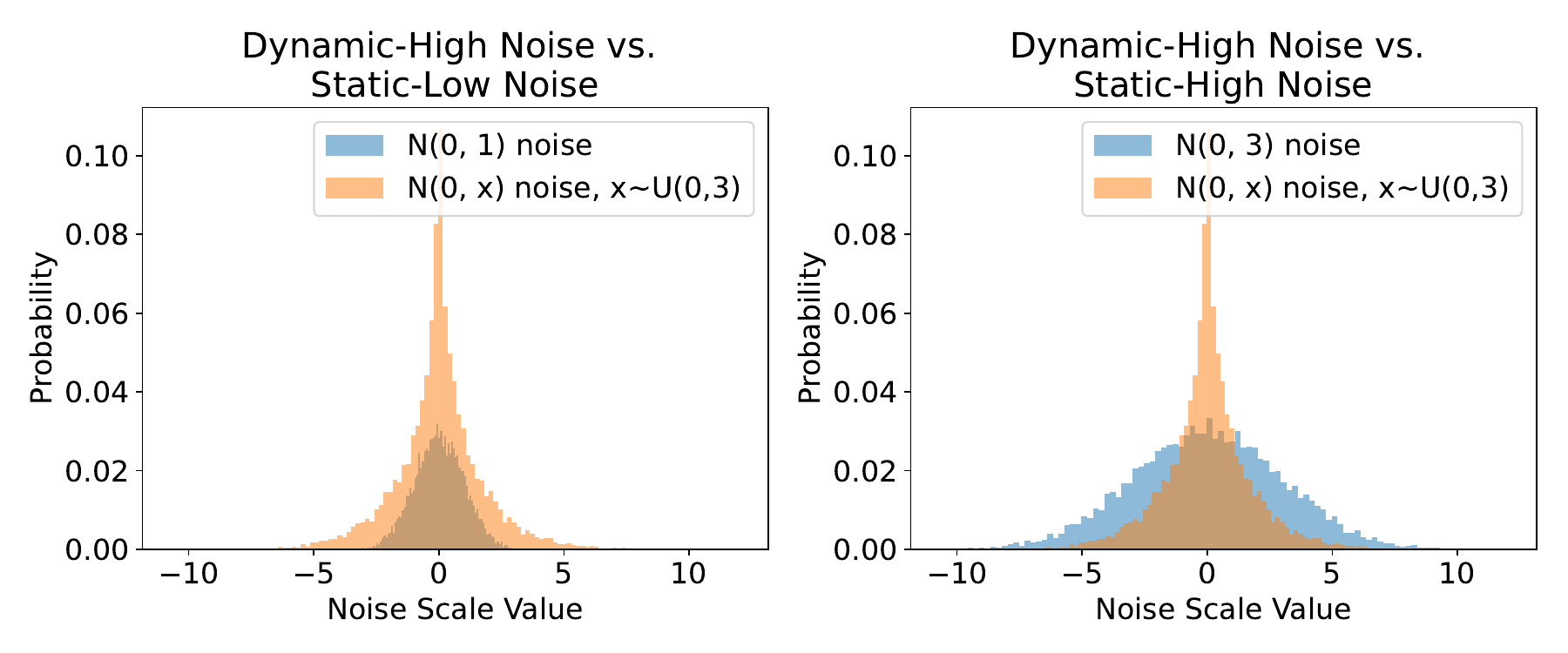}
    \caption{Comparison of noise scale distributions when sampled using static vs. dynamic noise scales. The dynamic noise scale sampling biases the distribution towards noise scale of lower values, but still maintain a good amount of noise scale of higher values.}
    \label{fig:dynamic_static_noise}
\end{figure}

\end{document}

%% file: math_command.tex
\usepackage{amsmath,amsfonts,bm}

\def\eqref#1{equation~\ref{#1}}

\def\1{\bm{1}}

\def\mA{{\bm{A}}}

\def\mD{{\bm{D}}}

\def\mF{{\bm{F}}}
\def\mG{{\bm{G}}}

\def\mI{{\bm{I}}}

\def\mP{{\bm{P}}}

\def\mX{{\bm{X}}}

\DeclareMathAlphabet{\mathsfit}{\encodingdefault}{\sfdefault}{m}{sl}
\SetMathAlphabet{\mathsfit}{bold}{\encodingdefault}{\sfdefault}{bx}{n}

\def\sD{{\mathbb{D}}}

\def\sR{{\mathbb{R}}}

\def\sZ{{\mathbb{Z}}}

